\documentclass[sn-mathphys-num]{sn-jnl}

\usepackage{graphicx}%
\usepackage{multirow}%
\usepackage{amsmath,amssymb,amsfonts}%
\usepackage{amsthm}%
\usepackage{mathrsfs}%
\usepackage[title]{appendix}%
\usepackage{xcolor}%
\usepackage{textcomp}%
\usepackage{manyfoot}%
\usepackage{booktabs}%
\usepackage{algorithm}%
\usepackage{algorithmicx}%
\usepackage{algpseudocode}%
\usepackage{listings}%

\theoremstyle{thmstyleone}%
\theoremstyle{thmstyletwo}%

\theoremstyle{thmstylethree}%

\begin{document}

\title[Article Title]{Advancing Human Action Recognition with Foundation Models trained on Unlabeled Public Videos}


\author*[1]{\fnm{Yang} \sur{Qian}}\email{qianyang@hawaii.edu}
\author[1]{\fnm{Yinan} \sur{Sun}}\email{sunyinan@hawaii.edu}
\author[1]{\fnm{Ali} \sur{Kargarandehkordi}}\email{kargaran@hawaii.edu}
\author[2]{\fnm{Parnian} \sur{Azizian}}\email{azizian@stanford.edu}
\author[2]{\fnm{Onur Cezmi} \sur{Mutlu}}\email{cezmi@stanford.edu}
\author[2]{\fnm{Saimourya} \sur{Surabhi}}\email{mourya.surabhi@stanford.edu}
\author[1]{\fnm{Zain} \sur{Jabbar}}\email{zjabbar@hawaii.edu}
\author[2]{\fnm{Dennis Paul} \sur{Wall}}\email{dpwall@stanford.edu}
\author[1]{\fnm{Peter} \sur{Washington}}\email{pyw@hawaii.edu}

\affil*[1]{\orgdiv{Information and Computer Sciences Department}, \orgname{University of Hawai’i at Mānoa}, \orgaddress{\city{Hawai’i}, \country{USA}}}
\affil[2]{\orgdiv{Stanford School of Medicine}, \orgname{Stanford University}, \orgaddress{\city{California}, \country{USA}}}


\abstract{The increasing variety and quantity of tagged multimedia content on a variety of online platforms offer a unique opportunity to advance the field of human action recognition. In this study, we utilize 283,582 unique, unlabeled TikTok video clips, categorized into 386 hashtags, to train a domain-specific foundation model for action recognition. We employ VideoMAE V2, an advanced model integrating Masked Autoencoders (MAE) with Vision Transformers (ViT), pre-trained on this diverse collection of unstructured videos. Our model, fine-tuned on established action recognition benchmarks such as UCF101 and HMDB51, achieves state-of-the-art results: 99.05\% on UCF101, 86.08\% on HMDB51, 85.51\% on Kinetics-400, and 74.27\% on Something-Something V2 using the ViT-giant backbone. These results highlight the potential of using unstructured and unlabeled videos as a valuable source of diverse and dynamic content for training foundation models. Our investigation confirms that while initial increases in pre-training data volume significantly enhance model performance, the gains diminish as the dataset size continues to expand. Our findings emphasize two critical axioms in self-supervised learning for computer vision: (1) additional pre-training data can yield diminishing benefits for some datasets and (2) quality is more important than quantity in self-supervised learning, especially when building foundation models.
}




\maketitle

\section{Introduction}\label{sec1}

Action recognition is a fundamental computer vision task that involves identifying and classifying human actions within video sequences. This technology is critically important in various applications, including security systems, interactive gaming, and healthcare monitoring. As social media platforms have grown, TikTok has become a prominent source of diverse human action videos. In response, we have developed a specialized foundation model trained on a carefully curated set of 283,582 TikTok video clips. This dataset, known as TikTokActions, represents a wide array of human activities and is designed to reflect real-world scenarios and cultural diversities. It includes unique, non-standard human actions, filling gaps in existing action recognition datasets and expanding the variety of human activities that current models can recognize.


Contrasting with the Kinetics dataset\cite{kay2017kinetics}, which also sources user-generated content but from YouTube, our TikTokActions dataset is composed exclusively of TikTok videos. These are characterized by their dynamic and spontaneous nature, often featuring rapid transitions and creative interactions that are less common in YouTube’s typically longer and more structured videos. This difference provides unique training challenges, enhancing the ability of models to interpret a broad spectrum of human actions in unpredictable environments.


We conducted a series of rigorous experiments to evaluate the efficacy of the TikTokActions dataset for pre-training computer vision models on action-related tasks. Using the advanced VideoMAE V2 model equipped with ViT-base and ViT-giant backbones\cite{wang2023videomae}, we explored the dataset's capabilities. The ViT-base provided a baseline for comparison against other models with similar architectures, while the ViT-giant was used to assess the maximum potential of our dataset. Subsequent fine-tuning on well-known datasets such as UCF101\cite{soomro2012ucf101}, HMDB51\cite{6126543}, Kinetics\cite{kay2017kinetics}, and Something-Something V2\cite{goyal2017somethingsomething} revealed that models pre-trained on TikTokActions often performed comparably to, or even better than, those trained solely on these traditional benchmarks or those pre-trained on significantly larger datasets, where enhancements tend to be incremental. 


Additionally, we investigated the relationship between the number of pre-training videos and model performance on downstream tasks. By training VideoMAE V2\cite{wang2023videomae} on increasing subsets of our collection, from 1,000 videos up to 6,000, we observed that performance remained robust, suggesting that massive datasets are not always essential for effective pre-training. This discovery emphasizes the efficiency of using online, unlabeled videos as a source for pre-training, challenging the prevailing reliance on large-scale video datasets for optimal model performance. Prior to deploying such methodologies in practical applications, we urge the research community to consider the ethical implications of using publicly available data for training deep learning models, particularly in terms of data scraping and privacy concerns.


\section{Related Work}\label{sec2}
\subsection{Human Action Recognition Datasets}\label{subsec1}

The evolution of action recognition research has been significantly shaped by the development and availability of benchmark datasets, which guide and validate advancements in computer vision methodologies. The UCF101\cite{soomro2012ucf101} dataset, known for its diverse array of human actions, has been instrumental in pioneering research within this domain. Likewise, the HMDB-51\cite{6126543} dataset, despite its smaller scale, offers a rich variety of human actions captured under realistic conditions, providing essential challenges for model evaluation.


Introduced in 2017, the Kinetics Human Action Video dataset \cite{kay2017kinetics} became a cornerstone in this field with over 400 distinct human action classes derived from YouTube videos. It set a new benchmark, catalyzing further research and development in action recognition technologies. Similarly, AVA \cite{gu2018ava} provides densely labeled video segments that allow for action localization in addition to recognition, offering a deeper understanding of context and action dynamics.


Emerging datasets such as Ego4D \cite{grauman2021ego4d}, which focuses on first-person video analysis, and HowTo100M \cite{miech2019howto100m}, which leverages instructional videos for action recognition, reflect the diversifying sources and modalities being explored. EPIC-Kitchens \cite{damen2018scaling}, ActivityNet \cite{caba2015activitynet}, and THUMOS \cite{idrees2017thumos} offer further variations in complexity and scenarios, pushing the envelope on what models can understand and predict.

Recent datasets have addressed limitations of earlier collections. The Something-Something dataset \cite{goyal2017somethingsomething} focuses on human-object interactions within everyday scenarios, offering a unique perspective that highlights the complexity of routine human activities. Moments in Time \cite{monfort2019moments} and Multi-Moment Dataset \cite{monfort2021multi} extend this by providing large-scale resources for temporal action understanding, capturing short videos that encompass a wide range of activities. The recent Tencent-MVSE dataset \cite{liu2022tencent}, another substantial contribution, is designed for multi-modal video similarity evaluation, adding another layer of complexity to action recognition tasks.


Building upon the diversity and scope of existing datasets, the TikTokActions dataset is meticulously designed to capture the rapidly evolving landscape of social media-driven human interactions. Unlike traditional datasets, which often focus on pre-defined, standard action categories, TikTokActions delves into a broader spectrum of spontaneous and culturally varied human behaviors, which are inherently more aligned with contemporary social media trends. This alignment is critical as platforms like TikTok foster unique user-generated content that is not only highly dynamic but also reflective of current societal norms and interactions.

In direct comparison to the Kinetics dataset, which largely aggregates content from YouTube, TikTokActions offers an array of actions driven by the unique constraints and creativity of TikTok's platform, such as shorter video lengths and a more diverse global user base. This distinction is significant as it introduces new challenges in action recognition, particularly in recognizing quick, context-driven actions that are less prevalent in the typically longer-form content of YouTube.

To substantiate the dataset's novelty and utility, we conducted a thorough analysis of the action categories present in TikTokActions compared to those in Kinetics and other major benchmarks. Our findings reveal that TikTokActions includes a higher proportion of non-standard actions, such as specific dance moves, challenge-based activities, and regional cultural expressions, which are underrepresented in other datasets. This inclusion not only enhances the dataset's coverage of real-world scenarios but also provides a more granular understanding of human actions within the digitally connected world.

By providing detailed statistics on the distribution of these categories, along with examples in our supplementary materials, we demonstrate the unique value of TikTokActions in filling the existing gaps within the field of action recognition. This detailed comparison and analysis affirm that TikTokActions is not just an addition to the plethora of action recognition datasets but a necessary evolution to keep pace with the changing dynamics of human activity in the digital age.


\subsection{Computer Vision Foundation Models}\label{subsec2}

Foundation models in computer vision have revolutionized the way we approach diverse tasks across modalities, offering scalable solutions that extend from image to video understanding. The Florence model \cite{yuan2021florence}, Specifically, encapsulates the progression of foundation models, adeptly offering transitions from broad scene outlines to detailed object-centric views and from static imagery to dynamic video sequences. Additionally, its effectiveness spans various modalities, from RGB imagery to textual data, establishing new standards in tasks like classification, object detection, retrieval, and particularly in action recognition.

Significant advancements have also been seen with the introduction of VideoMAE V2 \cite{wang2023videomae}. This self-supervised pre-training methodology stands out for its ability to train on datasets exceeding 1 million videos. With its innovative dual masking strategy and progressive training approach, VideoMAE V2 sets new standards in action recognition. Empirical evaluations demonstrate its superior performance across multiple benchmarks, emphasizing its effectiveness in real-world scenarios. 

Further developments include ViViT \cite{arnab2021vivit} and TimeSformer \cite{bertasius2021space}, which adapt transformer architectures specifically for video, offering improvements in handling the temporal dynamics of action sequences. These models demonstrate the potential of transformers in capturing complex, time-related patterns in video data.

The Quo Vadis model \cite{carreira2017quo} introduced a new approach by integrating the Kinetics dataset to set a performance benchmark, inspiring subsequent models to focus on improving temporal understanding and action recognition capabilities. Similarly, Temporal Relational Reasoning in videos \cite{zhou2018temporal} provided insights into the importance of temporal dynamics, influencing how subsequent models like Non-local Neural Networks \cite{wang2018non} handle long-range dependencies within video sequences.

A Closer Look at Spatiotemporal Convolutions \cite{tran2018closer} further explored the convolutional network architectures, focusing on how variations in spatial and temporal modules affect performance on action recognition tasks. This study paved the way for more specialized models like the Video Action Transformer Network \cite{girdhar2019video}, which introduced an approach to directly model relationships between actions and corresponding contextual features in videos.

Furthermore, 2D CNNs such as TSN \cite{wang2016temporal}, TSM \cite{lin2019tsm}, TANet \cite{liu2018tanet}, and TDN \cite{wang2021tdn} have been influential by effectively incorporating temporal information, a critical aspect of video understanding.

The use of 3D CNNs such as I3D \cite{carreira2017quo}, R(2+1)D \cite{tran2018closer}, ARTNet \cite{wang2018appearance}, and Slow-Fast \cite{feichtenhofer2019slowfast} has further pushed the envelope in spatial-temporal feature integration, enhancing model performance on complex video tasks.

In the domain of transformers, advancements have continued with models such as MViT \cite{fan2021multiscale} and Video Swin Transformer \cite{liu2021video} which have introduced ways to integrate spatial and temporal features more effectively, pushing the boundaries of action recognition performance. Additionally, UniFormer \cite{li2022uniformer} integrates local and global interactions more efficiently, which is crucial for detailed action understanding in videos.

The diversity of foundation models extends to multi-modal approaches, where models like Perceiver IO \cite{jaegle2021perceiver} and Data2Vec \cite{baevski2022data2vec} have been developed to handle various inputs ranging from text to video, showing flexibility and robustness in learning from heterogeneous data sources.

In the context of enhancing action recognition, SlowFast Networks \cite{feichtenhofer2019slowfast} and X3D \cite{feichtenhofer2020x3d} introduce novel architectural choices that optimize the processing of temporal variations and the physical intensity of actions. These developments underscore the trend towards more specialized and efficient models that are not only capable of high performance but also adaptability across different action recognition scenarios.

Our TikTokActions dataset, with its focus on spontaneous and diverse user-generated content, provides a rich testing ground for these advanced models. By evaluating these models on our dataset, we aim to explore how well they can adapt to and interpret the nuanced, quick-transition actions typical of content found on modern social media platforms. The insights gained can help tailor these foundation models to better handle real-world variability and complexity in human actions.

In our efforts to push the frontiers of action recognition further, we also draw upon the extensive research documented in the Multi-dataset Training of Transformers \cite{arnab2022multidataset}, which explores the robustness gained from training across varied video datasets. Such approaches are crucial as they contribute to the generalization capabilities of models, preparing them for deployment in diverse, unstructured environments.



\section{Dataset Construction}\label{sec3}

\subsection{Video Collection Overview}\label{subsec3}
To support our experimentation exploring the utility of unlabeled public data for pre-training foundation models for human action recognition, we curated an extensive collection of TikTok videos. This collection, consisting of over 280,000 video clips from 386 hashtags, captures a wide spectrum of human actions across various scenarios. The videos span diverse categories, from dance movements and fitness exercises to hand gestures and daily activities. Unlike traditional datasets like Kinetics, our collection uniquely incorporates elements of pop culture, memes, and rapid trend evolution, providing dynamic content that significantly boosts the model's ability to understand and generalize from real-world, culturally relevant scenarios.
\subsection{Video Selection Methodology}\label{subsec4}

To curate a diverse and relevant set of videos from TikTok, we initially examined several well-known action recognition datasets, including UCF101\cite{soomro2012ucf101}, HMDB51\cite{6126543}, and Google DeepMind Kinetics\cite{kay2017kinetics}. These datasets helped us to understand the broad categories of human actions typically represented in research. We identified four primary categories: dance, sports, fitness, and kinetics. 

Building upon these categories, we explored TikTok to identify relevant hashtags that capture a wide array of human actions within these domains. Our selection was strategically guided by six principles designed to ensure a comprehensive and culturally relevant collection: generalization, specification, noun substitution, noun-verb order, hashtag consolidation, and social/cultural relevance. These principles helped ensure that the videos not only cover a broad spectrum of actions but also include dynamic and culturally significant content such as memes, pop culture trends, and unique challenges—elements that are distinctly prevalent on TikTok. 

The social/cultural relevance was particularly important, allowing us to incorporate an additional category that includes a wide variety of actions originating from distinct social and cultural contexts, such as viral dance challenges and fitness trends. Popularity and engagement levels also played a crucial role in our selection process; we set a minimum threshold of 5 million views per hashtag to ensure that the content is not only relevant but also resonates widely with users, reflecting substantial public engagement and visibility. This approach allows us to harness rich, culturally nuanced content that can aid in training models to better understand and interpret human actions in real-world contexts.

\subsubsection{Generalization/Using the umbrella terms}\label{subsubsec1}
When comparing the hashtags we generated with those used by TikTok users, we observed that hashtags related to specific actions can often be represented in a very broad or inclusive manner on the platform. For instance, the hashtag \#basketball can serve as an umbrella term encompassing a wide array of hashtags, such as \#playingbasketball, \#dunking, \#dribbling.

\subsubsection{Specification}\label{subsubsec2}
There are situations where specificity is crucial. For instance, within the broader classification of Ballroom Dance, distinct dance styles such as Waltz, Quickstep, and Tango exist. Instead of solely using the hashtag \#ballroomdance, we include specific hashtags, including \#chacha, \#salsa, \#tango, and \#waltz, to ensure our datasets encompass pertinent human actions inherent to distinct dance styles.

\subsubsection{Noun Substitution}\label{subsubsec3}
While gathering short TikTok videos, we observed that a more effective approach for identifying the desired content is to employ a single noun instead of an entire verb phrase. For instance, the use of the hashtag \#piano suffices to compile a more extensive collection(see Table ~\ref{tab:hashtag_views}) of short videos instead of using \#pianoplaying. 

\begin{table*}[h]
\caption{Example of the expansion in data availability when removing nouns from the hashtag construction process.}
\centering
\begin{tabular}{|l|l|}
\hline
\textbf{Hashtags}     & \textbf{Number of Views} \\ \hline
\#piano               & 44.4 Billion                   \\ \hline
\#playingpiano  & 29.4 Million                   \\ \hline
\end{tabular}
\label{tab:hashtag_views}
\end{table*}

\begin{figure*}[t]
    \centering
    \includegraphics[width=\textwidth]{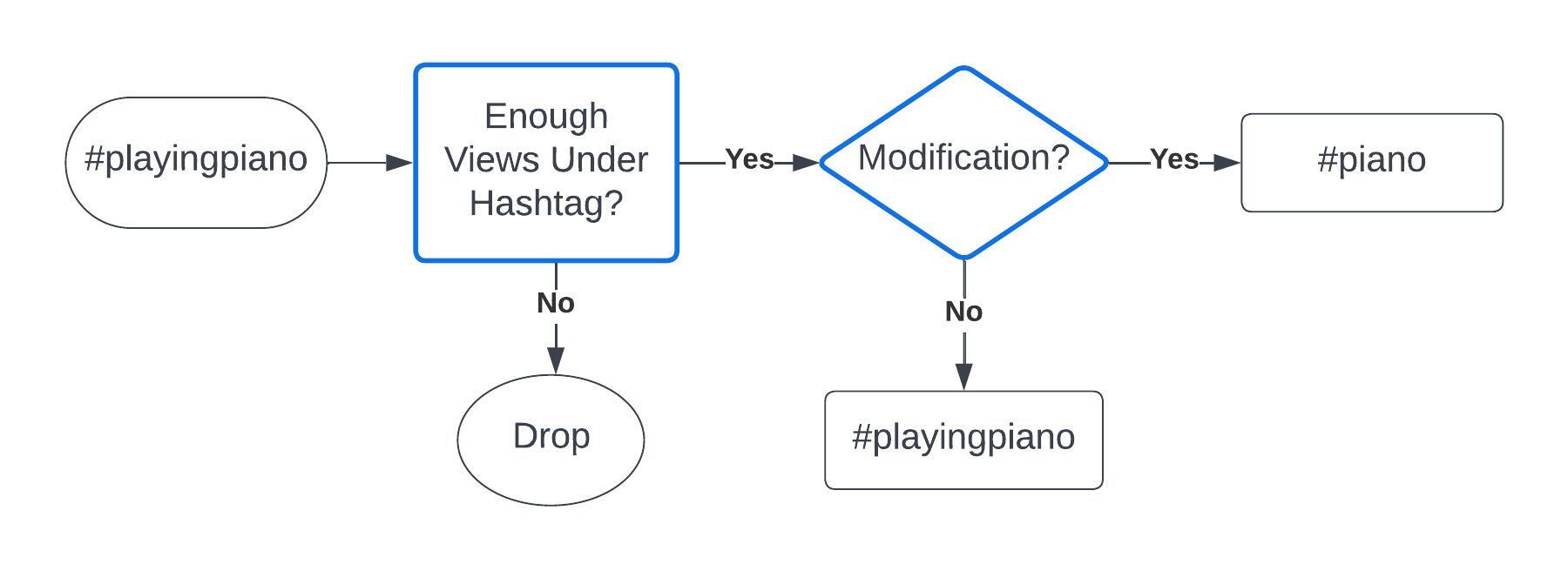}
    \caption{The hashtag selection and modification process.}
    \label{fig:hashtag_selection}
\end{figure*}

\subsubsection{Noun-Verb Order}
We noticed a preference among users for using noun phrases over verb phrases when describing their activities. For example, TikTok users frequently use hashtags like \#shoerepair instead of verb phrases like \#repairingshoes.

\subsubsection{Hashtag Consolidation}
When composing the hashtag list, we streamlined similar hashtags into a single one if they denote the same type of human action, prioritizing those with higher views on TikTok. For example, we retain \#cuttingcakes and remove similar hashtags like \#cutfruit, \#applecutting, and \#breadcutting.

\subsubsection{Social/cultural Considerations}
Culturally and socially relevant hashtags were included when crafting the hashtag list. For instance, the "flower challenge" originated as a dance challenge on TikTok initiated by a K-pop star, and TikTok users subsequently adopted the hashtag \#flowerchallenge when sharing their dance videos.

In Figure~\ref{fig:hashtag_selection}, we show an example flowchart of the hashtag selection and modification process.
Our final and complete list of hashtags along with video numbers is in Appendix A.

\subsection{Video Collection and Processing}
We curated a comprehensive video collection using the official TikTok research API, focusing on approximately 900 videos per hashtag from a set of hashtags identified through our selection process. Each video and its metadata were processed to facilitate model training, using tools such as \texttt{PySceneDetect} and \texttt{YOLO v8}\cite{reis2023realtime}. This processing resulted in a JSON file for each video, which included metadata capturing attributes like title, duration, and scene distribution, as analyzed by \texttt{PySceneDetect}.

\subsubsection{PySceneDetect and YOLOv8 Processing}

Our processing approach began with PySceneDetect, used to analyze the scene composition of each video. This tool helped identify distinct scenes within the videos, enabling us to focus on the longest scene clips, hypothesized to have the highest likelihood of containing significant human action.

Once the longest scenes were identified, the YOLOv8\cite{reis2023realtime} model was employed to detect human figures in these segments. The primary goal of this processing was to ensure accurate identification of videos containing human subjects, vital for the training of our models.

During the frame extraction phase, we used the ffprobe tool to determine the total frame count of each video clip. Due to the large number of frames in many videos, we applied subsampling for efficiency. Typically, a fixed number of frames (usually 10) were extracted at regular intervals throughout the video.

The extracted frames were then processed using the YOLOv8\cite{reis2023realtime} model to detect human presence. Frames that included at least one "person" label were counted towards the total number of frames containing humans for that particular video.

\begin{figure*}[t]
    \centering
    \includegraphics[width=\textwidth]{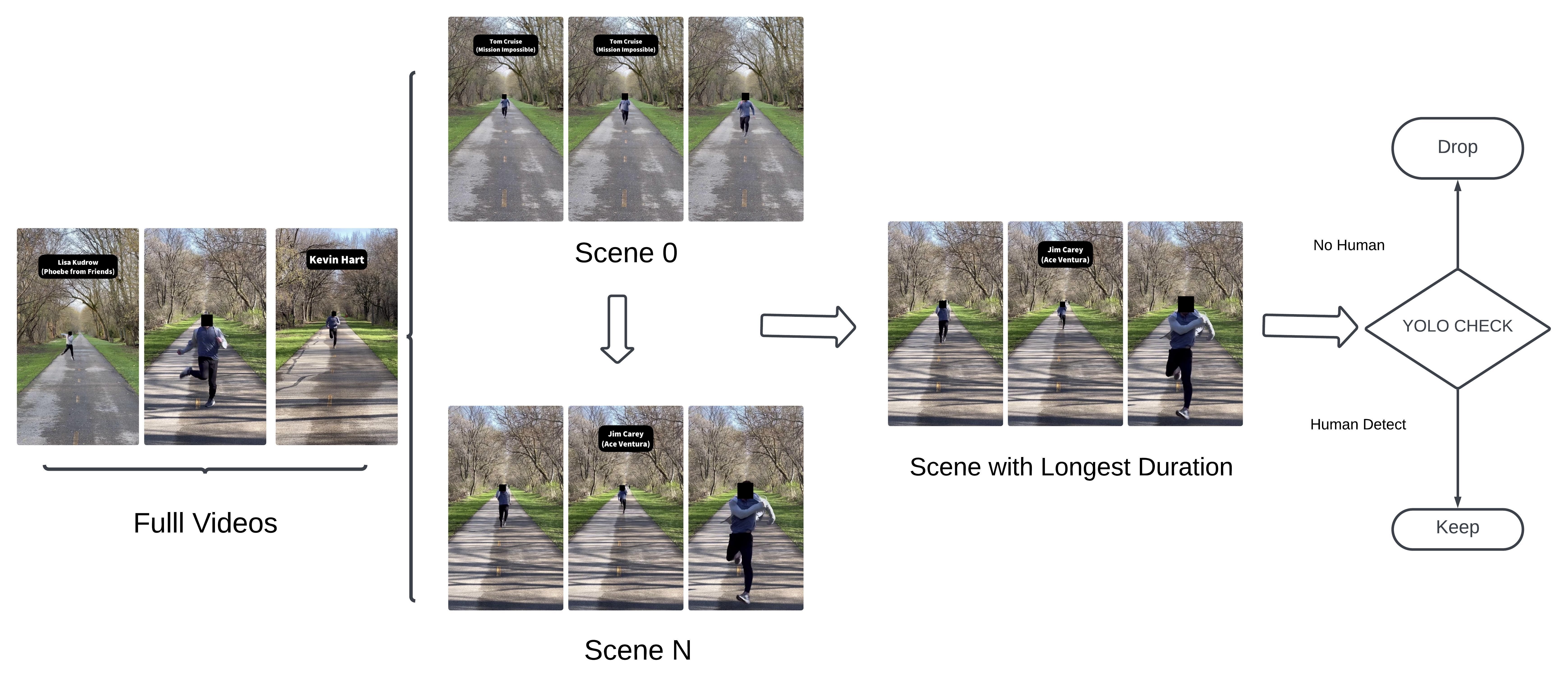}
    \caption{The clip extraction process from raw videos mined on TikTok.}
    \label{fig:clip}
\end{figure*}

Post-processing included applying a threshold for human presence; videos needed to exceed a pre-defined frame threshold with human detections to qualify. Videos that did not meet this threshold were subject to further review. This systematic approach helped compile aggregated statistics on the human presence in our video collection, ensuring the quality of content used for model training. The entire process, from video selection to frame extraction and human detection, was documented to maintain transparency and reproducibility. Figure ~\ref{fig:clip} illustrates the process of clip extraction from the raw videos sourced from TikTok.

\section{Performance on Benchmarks}\label{sec4}

In the domain of human action recognition, benchmarking performance is essential to understand the efficacy of a model. This section evaluates the performance of our model, initially pre-trained on the entire collection of 283,582 unlabeled TikTok video clips using VideoMAE V2\cite{wang2023videomae}. We employed both the ViT-base and ViT-giant backbones for pre-training to assess the impact of model capacity. Subsequently, the model was fine-tuned on Kinetics-400 (K400). Following this, the model underwent further fine-tuning on UCF-101\cite{soomro2012ucf101} and HMDB51\cite{6126543}, adopting a transfer learning approach akin to the experimental setups used in VideoMAE V2 studies. Additionally, we extended our evaluation to include the Something-Something V2\cite{goyal2017somethingsomething} dataset, assessing the model's adaptability and performance across varied action recognition contexts.

\subsection{Pre-training on VideoMAE V2}

VideoMAE V2\cite{wang2023videomae} is a state-of-the-art masked autoencoding framework, notable for its flexibility in backbone selection, ranging from base to billion-level configurations. We opted for both the ViT-base and ViT-giant backbones. The ViT-base was used to provide a baseline comparison with other models using similarly sized backbones, while the ViT-giant was employed to explore the maximum potential of the dataset, aiming to achieve the highest possible accuracy. The generalization capability of VideoMAE V2 pre-trained ViTs\cite{dosovitskiy2020image} as video foundation models has been demonstrated across multiple downstream tasks, highlighting its robustness and adaptability.

Our training was conducted on 8 NVIDIA A100 40GB GPUs, utilizing the full set of 283,582 unlabeled TikTok videos. The pre-training on the ViT-base backbone, which ran for 800 epochs and lasted for 20 days, provided a comprehensive base for developing a robust model capable of understanding a wide array of human actions. The ViT-giant backbone was pre-trained under similar conditions, running for 1200 epochs over 30 days, to push the performance boundaries further.

\subsection{Fine-Tuning on Established Datasets}

After initial pre-training, the model underwent two distinct paths of fine-tuning. The first path involved direct fine-tuning on the Kinetics-400 and Something-Something V2 datasets to assess immediate applicability to broad action recognition tasks. The results of this direct fine-tuning are detailed in Table~\ref{table:kinetics400_results} and Table~\ref{table:somethingv2_results}. The second path involved using the model fine-tuned on Kinetics-400 as an intermediary for transfer learning on UCF-101 and HMDB51, leveraging robust pre-training on TikTok videos. The outcomes of this transfer learning process are presented in Table~\ref{table:transfer_finetune_results}.

\begin{table*}[t]
\caption{Results on the Kinetics-400 dataset. We scale the pre-training of VideoMAE V2 to billion-level ViT-g model with curated TikTok clips. We report the fine-tuning accuracy of multiple view fusion (5×3). Both models are pre-trained and fine-tuned at the input of 16×224×224 and sampling stride $\tau$ = 4.}
\centering
\begin{tabular}{l l l l c c}
\toprule
\textbf{method} & \textbf{pre-train data} & \textbf{data size} & \textbf{epoch} & \textbf{ViT-B} & \textbf{ViT-g} \\ \midrule
MAE-ST \cite{feichtenhofer2022masked} & Kinetics400 & 0.24M & 1600 & 81.3 & - \\ 
VideoMAE V1 \cite{tong2022videomae} & Kinetics400 & 0.24M & 1600 & 81.5 & - \\ 
VideoMAE V2 \cite{wang2023videomae} & UnlabeledHybrid & 1.35M & 1200 & 81.5 & 87.2\\ 
VideoMAE V2 & TikTok & 0.28M & 800 & \textbf{81.1} & - \\ 
VideoMAE V2 & TikTok & 0.28M & 1200 & - & \textbf{85.51} \\ \midrule
\\ 
\bottomrule
\end{tabular}
\\
\label{table:kinetics400_results}
\end{table*}

\begin{table*}[t]
\caption{Results on the Something-Something V2 dataset. We scale the pre-training of VideoMAE V2 to billion-level ViT-g model with curated TikTok clips. We report the fine-tuning accuracy of multiple view fusion (2×3). Both models are pre-trained and fine-tuned at the input of 16×224×224 and sampling stride $\tau$ = 2.}
\centering
\begin{tabular}{l l l l c c}
\toprule
\textbf{method} & \textbf{pre-train data} & \textbf{data size} & \textbf{epoch} & \textbf{ViT-B} & \textbf{ViT-g} \\ \midrule
MAE-ST \cite{feichtenhofer2022masked} & Kinetics400 & 0.24M & 1600 & 70.8 & - \\ 
VideoMAE V1 \cite{tong2022videomae} & Sth-Sth V2 & 0.17M & 2400 & 70.8 & - \\ 
VideoMAE V2 \cite{wang2023videomae} & UnlabeledHybrid & 1.35M & 1200 & 71.2 & 77.0 \\ 
VideoMAE V2 & TikTok & 0.28M & 800 & \textbf{69.2} & - \\ 
VideoMAE V2 & TikTok & 0.28M & 1200 & - & \textbf{74.27} \\ \midrule
\\ 
\bottomrule
\end{tabular}
\\
\label{table:somethingv2_results}
\end{table*}

\begin{table*}[t]
\caption{Transfer Learning Results From K400 Fine-Tuning Through TikTok Pre-training}
\centering
\begin{tabular}{|c|c|c|}
\hline
\textbf{Dataset} & \textbf{ViT-b Top-1 (\%)} & \textbf{ViT-g Top-1 (\%)} \\
\hline
HMDB-51\cite{6126543} & 83.9 & 86.08 \\
UCF-101\cite{soomro2012ucf101} & 97.5 & 99.05 \\
\hline
\end{tabular}
\label{table:transfer_finetune_results}
\end{table*}

\textbf{Observations on data scaling}. We studied the influence of pre-training data size on the performance of VideoMAE V2. In this experiment, we pre-trained the video models with ViT-B and ViT-g backbones on our TikTok dataset, which contains approximately 0.28M videos. The fine-tuning accuracy is detailed in Table~\ref{table:kinetics400_results} for Kinetics-400 and Table~\ref{table:somethingv2_results} for Something-Something V2. Despite using approximately five times fewer pre-training samples compared to the UnlabeledHybrid dataset (1.35M videos), our model achieved comparable results. Specifically, the ViT-B backbone pre-trained on TikTok data achieved 81.1\% on Kinetics-400, which is only 0.4\% lower than the 81.5\% achieved using the much larger UnlabeledHybrid dataset. Similarly, for Something-Something V2, the ViT-B backbone pre-trained on TikTok data achieved 69.2\%, compared to 71.2\% with the UnlabeledHybrid dataset, showing a difference of 2\%. These small differences highlight the efficacy of our foundation model on TikTok videos, which did not include any information from the original fine-tuned dataset.

\textbf{Observations on model scaling}. We also explored how different model capacities affect performance. The fine-tuning results for models pre-trained with ViT-B and ViT-g backbones are presented in Table~\ref{table:kinetics400_results} and Table~\ref{table:somethingv2_results}. The ViT-g model, representing a billion-level parameter architecture, consistently demonstrated improved performance over the ViT-B model. For example, the ViT-g backbone pre-trained on TikTok data achieved 85.51\% on Kinetics-400, compared to 81.1\% for the ViT-B backbone, showing an improvement of 4.41\%. Similarly, for Something-Something V2, compared to the ViT-B backbone, there is an improvement of 5.07\%. These results suggest that increasing model capacity can further boost performance, validating the potential of large-scale pre-training on diverse video datasets like those from TikTok.

\textbf{Observations on transfer learning}. We further evaluated the performance of our pre-trained models by using Kinetics-400 fine-tuned models for transfer learning on HMDB-51 and UCF-101 datasets. The results are summarized in Table~\ref{table:transfer_finetune_results}. The ViT-B model achieved 83.9\% on HMDB-51 and 97.5\% on UCF-101, while the ViT-g model achieved 86.08\% on HMDB-51 and 99.05\% on UCF-101. 

Overall, our findings highlight the significant potential of using large, unlabeled video collections from platforms like TikTok for advancing human action recognition tasks. This approach capitalizes on the rich, varied nature of content available on social media, which, when used for pre-training, enhances the model's ability to generalize across different datasets and action recognition scenarios. The robust performance observed across both direct fine-tuning and transfer learning paths underscores the value of TikTok videos in training models for complex action recognition challenges. The TikTok dataset, which is unlabeled and lacks ground truth annotations, still produced promising results, demonstrating the model's capacity to learn effectively from unstructured video content.

\section{Dataset Size vs. Fine-Tuning Efficacy}\label{sec13}

To investigate the impact of pre-training data size on downstream model performance, we varied the size of the pre-training data from 1,000 to 6,000 videos, randomly selected from the TikTokActions dataset. Each data size level was used to pre-train the model in three independent runs to measure variance. Following pre-training, these models were fine-tuned on the UCF-101\cite{soomro2012ucf101} dataset. The primary metric for evaluating model performance was the accuracy observed on the test set of UCF-101, allowing us to draw conclusions about the relationship between pre-training data size and fine-tuning effectiveness.

\begin{figure}[htbp]
\centerline{\includegraphics[width=0.82\textwidth]{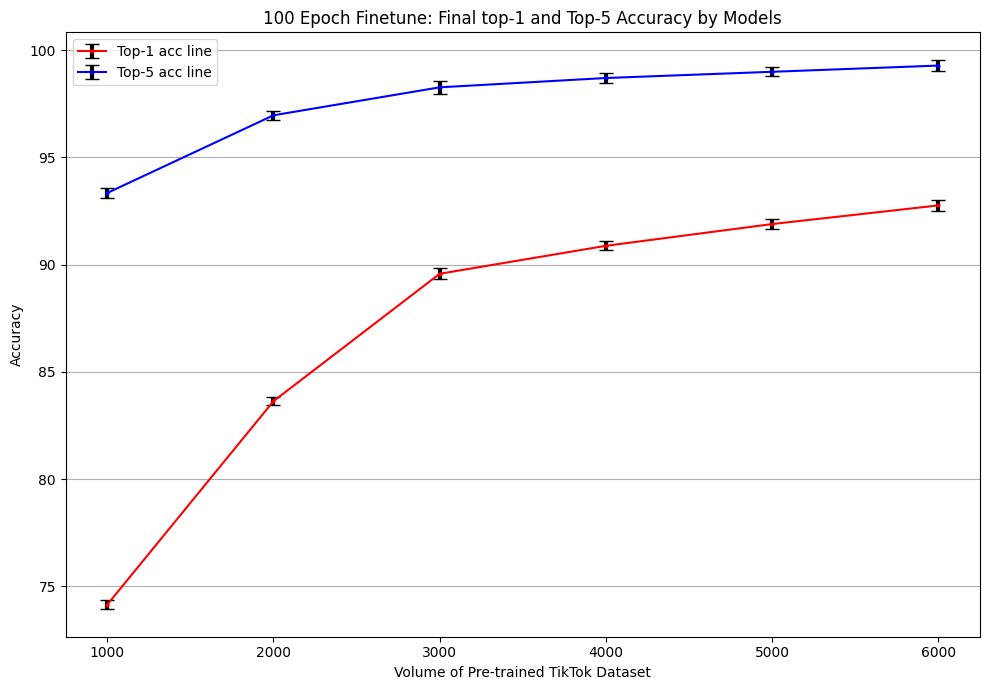}}
\caption{100 Epoch Finetune: Final top-1 and top-5 accuracy by amount of pre-training data. We demonstrate the impact of the volume of pre-trained TikTok dataset on the fine-tuning accuracy for both Top-1 and Top-5 metrics.}
\label{fig:accplot}
\end{figure}

From the result in Figure ~\ref{fig:accplot}, the volume of pre-trained data from the TikTokActions dataset was incrementally increased, and we observed a positive correlation with the Top-1 accuracy, particularly notable as the volume grew from 1,000 to 3,000 videos. However, subsequent increases in data volume yielded a reduced rate of improvement in Top-1 accuracy, indicating diminishing returns despite the larger dataset sizes. The Top-5 accuracy continued to show slight improvements, but the incremental benefit also appeared to diminish as the dataset size expanded beyond 3,000 videos. These findings suggest an asymptotic behavior in the benefit derived from increasing the volume of pre-training data for our models.

\section{Dataset Size vs. Fine-Tuning Efficacy}\label{sec14}

\begin{figure}[htbp]
\centerline{\includegraphics[width=0.82\textwidth]{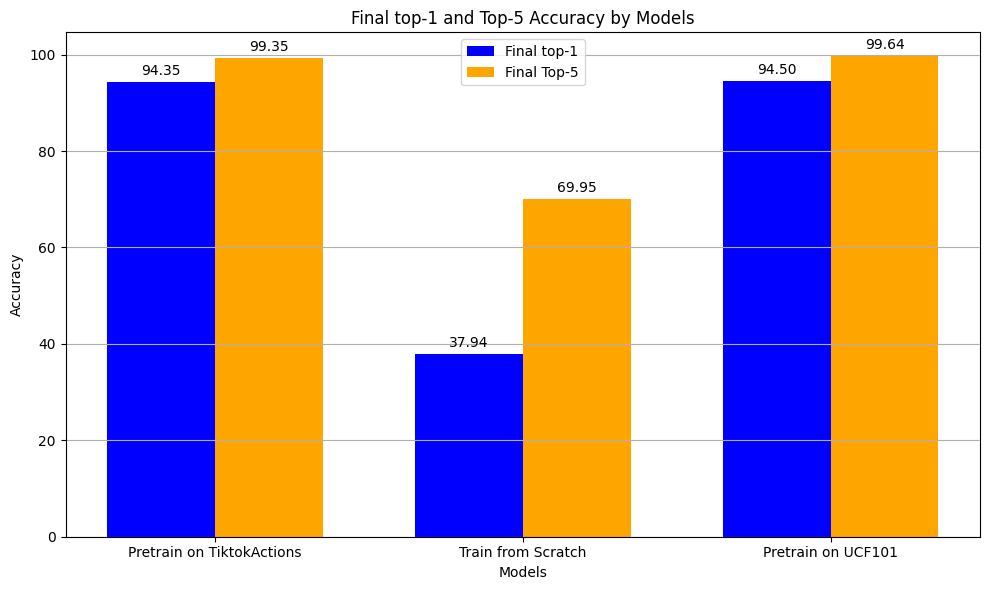}}
\caption{We compare the top-1 and top-5 accuracy of models pre-trained on TikTok actions, pre-trained from scratch, and pre-trained on UCF101 when fine-tuned to UCF101.}
\label{fig:tvu}
\end{figure}

We conducted further experiments to demonstrate the efficacy of the collection in pre-training models for action recognition. We used a targeted subset of 6,900 high-quality clips sourced from 73 TikTok hashtags. From Figure ~\ref{fig:tvu}, we can see the  pre-trained model achieved a top-1 accuracy of 94.35\% and a top-5 accuracy of 99.35\% when fine-tuning on the UCF101\cite{soomro2012ucf101}. This is remarkably close to the performance of models pre-trained on the UCF101 dataset itself (8155 from 101 classes), which exhibited a top-1 accuracy of 94.50\% and a top-5 accuracy of 99.64\%. The comparable accuracies highlight the mined dataset's quality and its potential to serve as a substantial pre-training resource, even with fewer and more focused training samples. The model trained from scratch, without the benefit of such a rich dataset, performed significantly poorer, reinforcing the value of pre-training on high-quality, domain-specific data. This outcome attests to the robustness and generalization capabilities of models pre-trained on the curated dataset.

\section{Discussion}

The curation of TikTok videos for enhancing human action recognition models provides a unique contribution to the field, especially by incorporating a variety of distinct yet non-traditional human actions. These actions, sourced from the TikTok platform, were carefully selected to complement the more conventional actions typically found in established datasets like UCF101\cite{soomro2012ucf101} and HMDB51\cite{6126543}. This approach not only diversifies the types of human actions available for model training but also qualitatively enriches the dataset, thereby challenging and extending the adaptability of action recognition algorithms.

Integrating these unique and diverse actions from TikTok, we have expanded the range of actions that machine learning models can recognize, addressing the dynamic nature of human actions in digital media where new and unconventional actions frequently emerge. This strategy highlights the value of leveraging culturally rich and dynamic user-generated content as a training resource in an era dominated by digital media.

Furthermore, our findings challenge the conventional wisdom that larger datasets always yield better pre-training results. The competitive performance of our models, initially pre-trained on a curated subset of TikTok videos and subsequently fine-tuned on Kinetics-400\cite{kay2017kinetics} (K400) and Something-Something V2(Sth-Sth)\cite{goyal2017somethingsomething}, suggests that a well-curated, smaller dataset can sometimes outperform a larger, more generic one. This observation encourages a shift towards more efficient dataset compilation and model training paradigms, emphasizing data quality over quantity.

Looking ahead, exploring the potential of using the full breadth of the TikTok video collection for pre-training could further enhance model performance and generalization across various tasks. Additionally, integrating these videos with other datasets, such as Kinetics-400 and Something-Something, could provide deeper insights into the unique characteristics and potential synergies between different types of video content.

An emerging area of interest is the application of weekly self-supervised learning techniques, which involve regular, unsupervised updates to the model based on new data inflows. Such techniques could significantly boost the adaptability and accuracy of models trained with dynamic, culturally relevant content from platforms like TikTok.

\section{Ethics}

This study has been approved under the University of Hawaii Institutional Review Board (IRB) under protocol number 2024-00396. Furthermore, this research adheres to strict ethical guidelines regarding the use of online video data, specifically videos collected from TikTok. All videos were legally accessed via the TikTok Research API, and our data collection process is in full compliance with the TikTok Research API Terms of Service. Importantly, only those videos where the user granted explicit download permission were included in our training process. Equally as importantly, we deleted all of the data after model training as an additional layer of security.

We place a high priority on user privacy. To protect individual privacy, we have implemented face obfuscation techniques in the video clips included in our paper. Moreover, in line with the terms of service, we do not distribute any TikTok user content directly. Instead, we focus on training a foundation model, which we only use to identify action patterns, not personal biometric features.

\textbf{Genderal Ethical Considerations for Online Video Research:} The use of online video data poses unique challenges and ethical considerations, particularly regarding privacy, informed consent, and the potential risks associated with the use of such data. Our study emphasizes the importance of adhering to ethical principles that safeguard participant interests. We have made significant efforts to minimize potential harm by anonymizing the data and not sharing the curated dataset. We recommend that researchers adhere to such privacy protections.

Furthermore, we advocate for the inclusion of clearer guidelines in the terms of service on video platforms, which would allow users of the platforms to explicitly exclude their videos from research use. This approach does not replace the need for informed consent but is a step towards enhancing ethical practices in online video research. We aim to foster a transparent and informed discussion about the use of online video data in computational research.

We also argue that the use of the models plays a critical role in deciding the ethical standards that are required. Predicting human actions such as running, swimming, or playing piano from videos is a common task that is unlikely to lead to harm, especially given how common action recognition computer vision research is. By contrast, if the AI were used to predict sensitive properties about the end user such as mental health diagnoses, income, political party, or anything else along those lines, then the standard for explicit consent and ethical practices would rise to a much higher level \cite{jmirEthicsPaper}.

\backmatter

\bmhead{Acknowledgements}

This work was made possible by a multitude of supporters and funding sources. We thank the National Science Foundation for their generous support through the CyberTraining grant (Award No. 2118222) under the HI-DSI project. The technical assistance and advanced computing resources provided by the University of Hawaii Information Technology Services--Cyberinfrastructure, partially funded by NSF CC* awards \#2201428 and \#2232862 (Koa), have been invaluable. The development and acceleration of cloud computing through JetStream2 have significantly contributed to our research; we acknowledge David Y. Hancock, Jeremy Fischer, John Michael Lowe, Winona Snapp-Childs, Marlon Pierce, Suresh Marru, J. Eric Coulter, Matthew Vaughn, Brian Beck, Nirav Merchant, Edwin Skidmore, and Gwen Jacobs for this computational infrastructure (\href{https://doi.org/10.1145/3437359.3465565}{DOI: 10.1145/3437359.3465565}). The AWS Cloud Credit for Research program has also played a crucial role in providing computational resources for our work. We used ChatGPT to edit the grammar of our manuscript and to re-phrase sentences that were originally worded unclearly. However, all contents in this manuscript are original ideas and analyses conducted by the authors.

\begin{appendices}

\section{Sample Clips}

In the appendix, we will present additional insights into our dataset, specifically focusing on visual and statistical aspects. This includes select screenshots of our video clips, which exemplify the diversity and context of the actions captured within our dataset. Alongside these visuals, we will include the figure, which provides the number distributions of videos for each hashtag. Furthermore, we will present the distribution of video durations, offering a comprehensive view of the temporal characteristics of our clips. These elements combined will give a more complete picture of the dataset's richness.

\begin{figure*}[t]
    \centering
    \includegraphics[width=\textwidth]{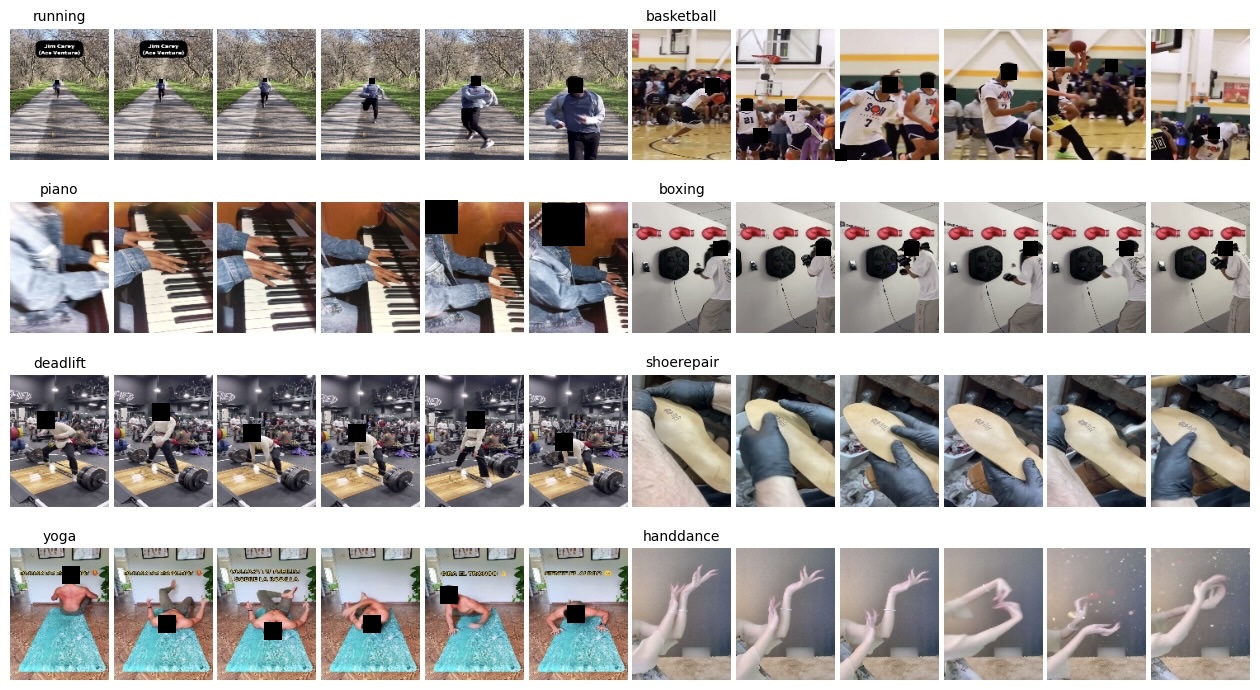}
    \caption{Example clips from the TikTokActions dataset.}
    \label{fig:clipsample}
\end{figure*}

In Figure~\ref{fig:clipsample}, we present the extensive diversity of the TikTokActions dataset, which captures a wide range of human activities tagged across multiple domains such as fitness, sports, and dance. The collection features a variety of video types that include full-body human actions, clips focusing on hand movements, and scenarios involving multiple individuals. 
The presence of such varied content provides a rich learning ground for self-supervised learning models to extract and learn a comprehensive set of features from human actions, gestures, and interactions, aiding in the enhancement of action recognition capabilities.

\section{Dataset Composition}
\begin{figure}[H]
    \centering
    \includegraphics[width=\textwidth]{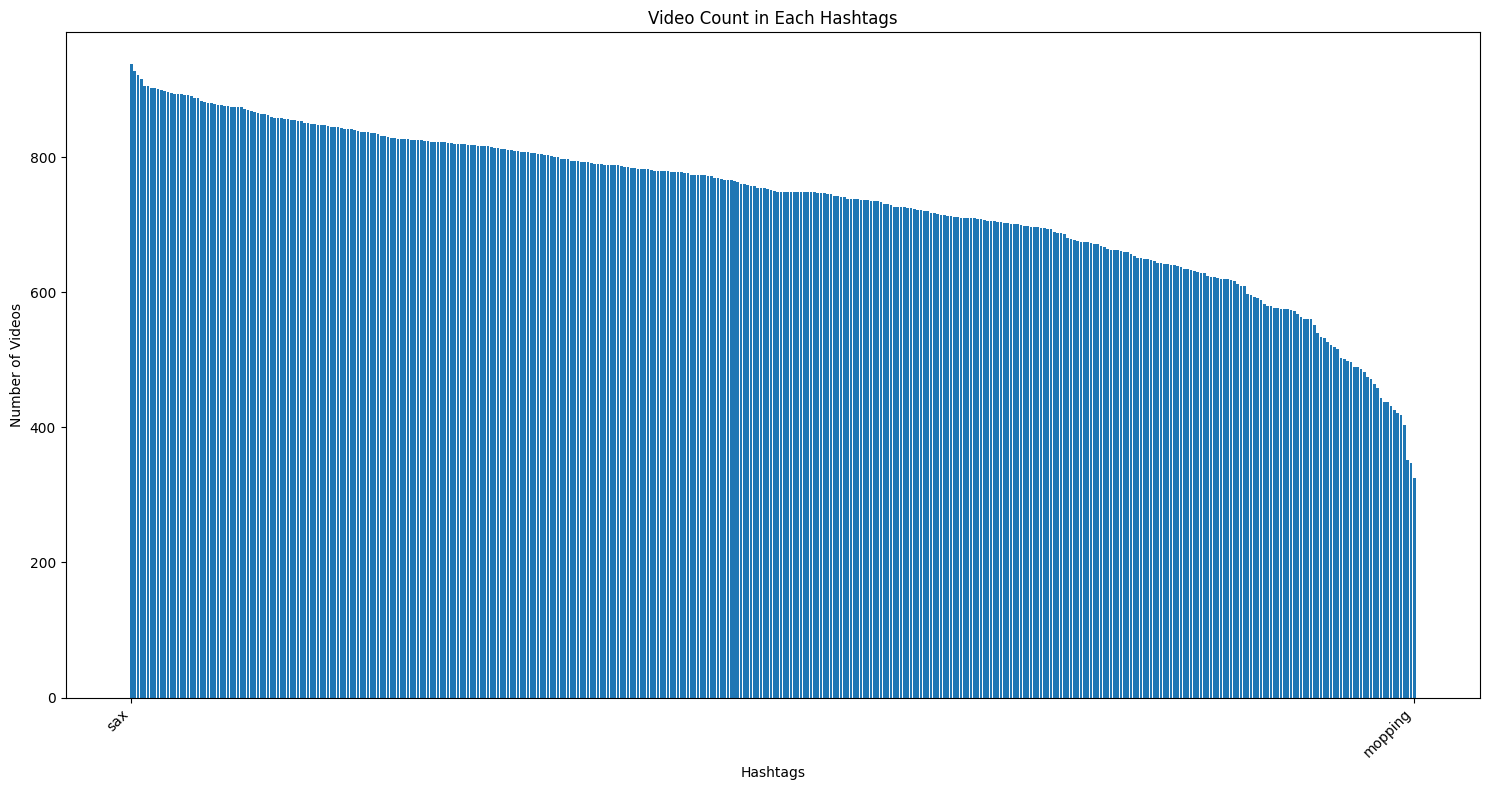}
    \caption{Bar chart showing the number of videos per hashtag in our dataset, 
    demonstrating the frequency and distribution of action categories.}
    \label{fig:bar_chart}
\end{figure}

Our dataset offers a diverse array of video counts across different hashtags. Figure~\ref{fig:bar_chart} illustrates the distribution of videos across various hashtags, indicating the prevalence of certain actions within our dataset. On average, each hashtag is associated with approximately 735 videos. The 'sax' category stands out with the maximum count of 938 videos, while the 'mopping' category has the minimum, with 325 videos. 

\begin{figure}[H]
    \centering
    \includegraphics[width=0.8\textwidth]{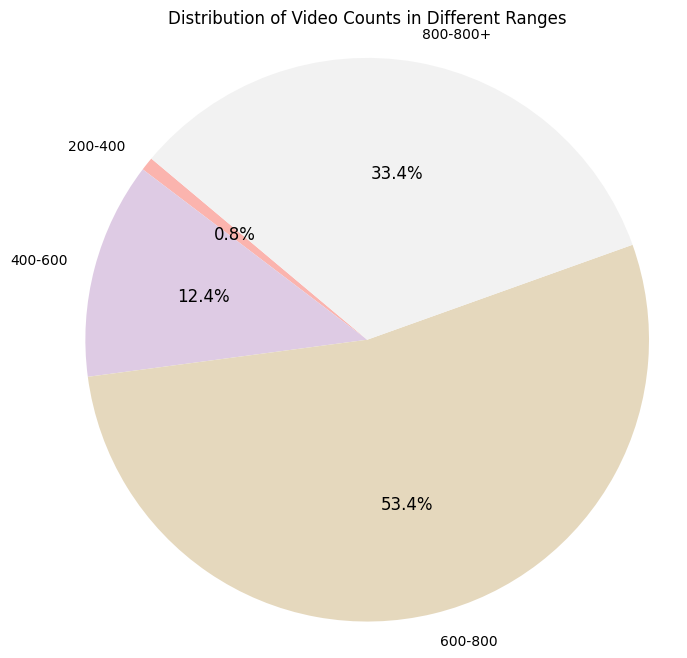}
    \caption{Video counts in different ranges, 
    highlighting the dataset's variety and the volume of data available for model training.}
    \label{fig:pie_chart}
\end{figure}

Conversely, Figure~\ref{fig:pie_chart} displays a pie chart that categorizes the dataset into four video count ranges: 200-400, 400-600, 600-800, and 800+ videos per category. It is noteworthy that a significant portion, 53.4\%, of the hashtags have a video count that falls within the 600-800 range.

\section{Dataset Duration}
\label{sec:dataset_duration}
In the composition of the TikTokActions dataset, a key focus was on ensuring that each video clip is sufficiently long to capture a complete action, while also maintaining a manageable length for analysis. To this end, we established a minimum clip duration of 3.5 seconds, ensuring each clip is long enough to encompass a full action. Simultaneously, to avoid excessively long clips, which can complicate analysis and may contain extraneous information, we capped the maximum duration at 10 seconds. For videos exceeding this length, only the most relevant 10-second segment, typically featuring the most significant action, was selected. This methodology results in a dataset that is both comprehensive and tailored for efficient action recognition analysis.

The following table details the distribution of video durations within our dataset, post these adjustments:

\begin{table*}[t]
\caption{Distribution of video durations in the TikTokActions dataset. The data reflects the adherence to the minimum and maximum duration thresholds for clip selection.}
\centering
\begin{tabular}{|c|c|}
\hline
\textbf{Duration Range (seconds)} & \textbf{Number of Videos} \\ \hline
3.5-5 & 30103 \\ \hline
5-10 & 81356 \\ \hline
10+ & 172123 \\ \hline
\end{tabular}
\label{tab:video_duration}
\end{table*}




\end{appendices}


\bibliography{sn-bibliography}

\end{document}